\documentclass[10pt,twocolumn,letterpaper]{article}

\usepackage{cvpr}
\usepackage{times}
\usepackage{epsfig}
\usepackage{graphicx,epstopdf}
\usepackage{amsmath}
\usepackage{amssymb}
\usepackage{authblk}
\usepackage{placeins}
\newcommand{\bigcell}[2]{\begin{tabular}{@{}#1@{}}#2\end{tabular}}

\newcommand{\Tref}[1]{Table~\ref{#1}}
\newcommand{\Eref}[1]{Eq.~(\ref{#1})}
\newcommand{\Fref}[1]{Fig.~\ref{#1}}
\newcommand{\Sref}[1]{Sec.~\ref{#1}}

\cvprfinalcopy 

\def\cvprPaperID{****} 

\begin{document}

\title{Fisher Kernel for Deep Neural Activations}

\author{Donggeun Yoo}
\author{Sunggyun Park}
\author{Joon-Young Lee}
\author{In So Kweon}
\affil{KAIST\\Daejeon, 305-701, Korea.\\\tt\small dgyoo@rcv.kaist.ac.kr, sunggyun@kaist.ac.kr, jylee@rcv.kaist.ac.kr, iskweon77@kaist.ac.kr}


\maketitle

\begin{abstract}
Compared to image representation based on low-level local descriptors, deep neural activations of Convolutional Neural Networks (CNNs) are richer in mid-level representation, but poorer in geometric invariance properties.
In this paper, we present a straightforward framework for better image representation by combining the two approaches.
To take advantages of both representations, we propose an efficient method to extract a fair amount of multi-scale dense local activations from a pre-trained CNN. We then aggregate the activations by Fisher kernel framework, which has been modified with a simple scale-wise normalization essential to make it suitable for CNN activations.
Replacing the direct use of a single activation vector with our representation demonstrates significant performance improvements: +17.76 (Acc.) on MIT Indoor 67 and +7.18 (mAP) on PASCAL VOC 2007. The results suggest that our proposal can be used as a primary image representation for better performances in visual recognition tasks.
\end{abstract}

\vspace{-4mm}
\section{Introduction}

\begin{figure*}[t]
\begin{center}
\includegraphics[width=1\linewidth]{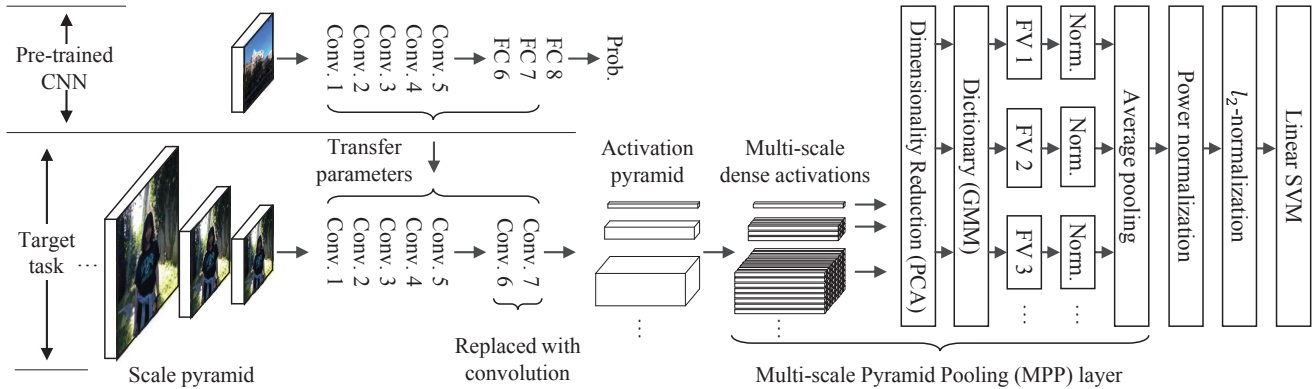}
\end{center}
   \caption{A pipeline of the proposed method. Given a pre-trained CNN, we replace the first two fully connected layers with the two equivalent convolutional layers to efficiently obtain large amount of multi-scale dense activations. The activations are followed by the Multi-scale Pyramid Pooling (MPP) layer we suggest. The consequent image representation is combined with the linear SVM for the target classification task.}
\label{FIG_PIPELINE}
\end{figure*}

Image representation is one of the most important factors that affect performance on visual recognition tasks. Barbu~\etal~\cite{FMRI} introduced an interesting experiment that a simple classifier along with human brain-scan data substantially outperforms the state-of-the-art methods in recognizing action from video clips.

With a success of local descriptors~\cite{SIFT}, many researches devoted deep study to global image representation based on a Bag-of-Word (BOW) model~\cite{VGOOGLE} that aggregates abundant local statistics captured by hand-designed local descriptors. The BOW representation is further improved with VLAD~\cite{ORGVLAD} and Fisher kernel~\cite{IFK, FK} by adding higher order statistics. One major benefit of these global representations based on local descriptors is their invariance property to scale changes, location changes, occlusions and background clutters. 

In recent computer vision researches, drastic advances of visual recognition are achieved by deep convolutional neural networks (CNNs) \cite{LECUN}, which jointly learn the whole feature hierarchies starting from image pixels to the final class posterior with stacked non-linear processing layers. A deep representation is quite efficient since its intermediate templates are reused. However, the deep CNN is non-linear and have millions of parameters to be estimated. It requires strong computing power for the optimization and large training data to be generalized well. The recent presence of large scale ImageNet \cite{IMAGENET} database and the raise of parallel computing contribute to the breakthrough in visual recognition. Krizhevsky~\etal~\cite{ALEX} achieved an impressive result using a CNN in large-scale image classification.

Instead of training a CNN for a specific task, intermediate activations extracted from a CNN pre-trained on independent large data have been successfully applied as a generic image representation. Combining the CNN activations with a classifier has shown impressive performance in wide visual recognition tasks such as object classification \cite{OFFTHESHELF, DECAF, SIVIC, SPP, DEVIL}, object detection \cite{RCNN, SPP}, scene classification \cite{OFFTHESHELF, VLAD, MITPLACE}, fine-grained classification \cite{OFFTHESHELF, FINERCNN}, attribute recognition \cite{PANDA}, image retrieval \cite{NEURALCODE}, and domain transfer \cite{DECAF}.

For utilizing CNN activations as a generic image representation, a straightforward way is to extract the responses from the first or second fully connected layer of a pre-trained CNN by feeding an image and to represent the image with the responses~\cite{DECAF, NEURALCODE, SPP, RCNN}. However, this representation is vulnerable to geometric variations. There are techniques to address the problem. A common practice is exploiting multiple jitterred images (random crops and flips) for data augmentation. Though the data augmentation has been used to prevent over-fitting \cite{ALEX}, recent researches show that \textit{average pooling}, augmenting data and averaging the multiple activation vectors in a test stage, also helps to achieve better geometric invariance of CNNs while improving classification performance by +2.92\% in \cite{DEVIL} and +3.3\% in \cite{OFFTHESHELF} on PASCAL VOC 2007.

A different experiment for enhancing the geometric invariance on CNN activations was also presented. Gong~\etal~\cite{VLAD} proposed a method to exploit multi-scale CNN activations in order to achieve geometric invariance characteristic while improving recognition accuracy. They extracted dense local patches at three different scales and fed each local patch into a pre-trained CNN. The CNN activations are aggregated at finer scales via VLAD encoding which was introduced in ~\cite{ORGVLAD}, and then the encoded activations are concatenated as a single vector to obtain the final representation.

In this paper, we introduce a \textit{multi-scale pyramid pooling} to improve the discriminative power of CNN activations robust to geometric variations. A pipeline of the proposed method is illustrated in Figure \ref{FIG_PIPELINE}. Similar to \cite{VLAD}, we also utilize multi-scale CNN activations, but present a different pooling method that shows better performance in our experiments. Specifically, we suggest an efficient way to obtain abundant amount of multi-scale local activations from a CNN, and aggregate them using the state-of-the-art Fisher kernel~\cite{IFK, FK} with a simple but important scale-wise normalization, so called \textit{multi-scale pyramid pooling}. Our proposal demonstrates substantial improvements on both scene and object classification tasks compared to the previous representations including a single activation, the average pooling~\cite{OFFTHESHELF, DEVIL}, and the VLAD of activations~\cite{VLAD}. Also, we demonstrate object confidence maps which is useful for object detection/localization though only category-level labels without specific object bounding boxes are used in training.

According to our empirical observations, replacing a VLAD kernel with a Fisher kernel does not present significant impact, however it shows meaningful performance improvements when our pooling mechanism that takes an average pooling after scale-wise normalization is applied. It implies that the performance improvement of our representation does not come just from the superiority of Fisher kernel but from the careful consideration of neural activation's property dependent on scales.

\begin{figure*}
\begin{center}
\setlength{\tabcolsep}{1.7pt}
\includegraphics[width=1\linewidth]{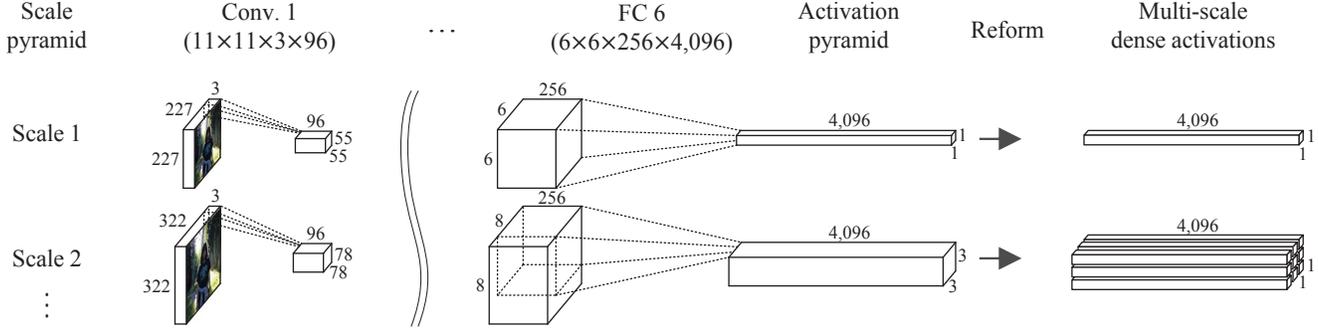}
\end{center}
\vspace{-3mm}
\caption{Obtaining multi-scale local activations densely from a pre-trained CNN. In this figure, the target layer is the first fully connected layer (FC6). Because FC6 can be equally implemented by a convolutional layer containing 4,096 filters of 6$\times$6$\times$256 size, we can obtain an activation map where spatial ordering of local descriptors is conserved. A single pre-trained CNN is shared for all scales.}
\vspace{-3mm}
\label{FIG_LOCALDESC}
\end{figure*}

\section{Multi-scale Pyramid Pooling}
In this section, we first review the Fisher kernel framework and then introduce a \textit{multi-scale pyramid pooling} which adds a Fisher kernel based pooling layer on top of a pre-trained CNN.

\subsection{Fisher Kernel Review}
\label{FISHER_REVIEW}
The Fisher kernel framework on a visual vocabulary is proposed by Perronnin~\etal in \cite{FK}. It extends the conventional Bag-of-Words model to a probabilistic generative model. It models the distribution of low-level descriptors using a Gaussian Mixture Model (GMM) and represents an image by considering the gradient with respect to the model parameters. Although the number of local descriptors varies across images, the consequent Fisher vector has a fixed-length, therefore it is possible to use discriminative classifiers such as a linear SVM. 

Let $\mathbf{x}$ denote	 a $d$-dimensional local descriptor and $\mathbf{G_\lambda} = \{\mathbf{g}_k,  k \! = \! 1...K\}$ denote a pre-trained GMM with $K$ Gaussians where $\mathbf{\lambda}=\{\omega_k,\mu_k,\sigma_k, k \! = \! 1...K\}$. For each visual word $\mathbf{g}_k$, two gradient vectors, $\mathcal{G}_{\mu_k}\in\Re^d$ and $\mathcal{G}_{\sigma_k}\in\Re^d$, are computed by aggregating the gradients of the local descriptors extracted from an image with respect to the mean and the standard deviation of the $k^\text{th}$ Gaussian. Then, the final image representation, \textit{Fisher vector}, is obtained by concatenating all the gradient vectors. Accordingly, the Fisher kernel framework represents an image with a $2Kd$-dimensional Fisher vector $\mathcal{G}\in\Re^{2Kd}$.

Intuitively, a Fisher vector includes the information about directions of model parameters to best fit the local descriptors of an image to the GMM. The fisher kernel framework is further improved in \cite{IFK} by the additional two-stage normalizations: power-normalization with the factor of 0.5 followed by $\ell_2$-normalization. Refer to \cite{IFK} for the theoretical proofs and details. 


\subsection{Dense CNN Activations}

To obtain multi-scale activations from a CNN without modification, previous approach cropped local patches and fed the patches into a network after resizing the patches to the fixed size of CNN input. However, when we extract multi-scale local activations densely, the approach is quite inefficient since many redundant operations are performed in convolutional layers for overlapped regions. 

To extract dense CNN activations without redundant operations, we simply replace the fully connected layers of an existing CNN with equivalent multiple convolution filters along spatial axises.
When an image larger than the fixed size is fed, the modified network outputs multiple activation vectors where each vector is CNN activations from the corresponding local patch. 
The procedure is illustrated in \Fref{FIG_LOCALDESC}. With this method, thousands of dense local activations (4,410 per image) from multiple scale levels are extracted in a reasonable extraction time (0.46 seconds per image) as shown in \Tref{TAB_TIME}.

\begin{table}[t]
\setlength{\tabcolsep}{1.6pt}
\small
\begin{center}
\setlength{\tabcolsep}{1mm}
\begin{tabular}{|l|c|c|c|c|}\hline
Image scales&1$\sim$4&1$\sim$5&1$\sim$6&1$\sim$7\\
Number of activations&270&754&1,910&4,410\\\hline\hline
Naive extraction (sec)&1.702&4.941&11.41&27.64\\
Proposed extraction (sec)&\textbf{0.0769}&\textbf{0.1265}&\textbf{0.2420}&\textbf{0.4635}\\\hline
\end{tabular}
\end{center}
\caption{Average time for extracting multi-scale dense activations per image. With Caffe reference model \cite{CAFFE}, FC7 activations are extracted from 100 random images of PASCAL VOC 2007. All timings are based on a server with a CPU of 2.6GHz Intel Xeon and a GPU of GTX TITAN Black.}
\label{TAB_TIME}
\end{table}

\subsection{Multi-scale Pyramid Pooling (MPP)}
For representing an image, we first generate a scale pyramid for the input image where the minimum scale image has a fixed size of a CNN and each scale image has two times larger resolution than the previous scale image. We feed all the scaled images into a pre-trained CNN and extract dense CNN activation vectors. Then, all the activation vectors are merged into a single vector by our multi-scale pyramid pooling.

If we consider each activation vector as a local descriptor, it is straightforward to aggregate all the local activations into a Fisher vector as explained in \Sref{FISHER_REVIEW}.
However, CNN activations have different scale properties compared to SIFT-like local descriptors, as will be explained in \Sref{ANALYSIS_SCALE}. To adopt the Fisher kernel suitable to CNN activation characteristics, we introduce adding a \textit{multi-scale pyramid pooling layer} on top of the modified CNN as follows.

Given a scale pyramid $S$ containing $N$ scaled image and local activation vectors $\mathbf{x}_s$ extracted from each scale $s\in S$, we first apply PCA to reduce the dimension of activation vectors and obtain $\mathbf{x}'_s$. Then, we aggregate the local activation vectors $\mathbf{x}'_s$ of each scale $s$ to each Fisher vector $\mathcal{G}^{s}$. After Fisher encoding, we have $N$ Fisher vectors and they are merged into one global vector by average pooling after $\ell_2$-normalization as
\begin{equation}
\mathcal{G}^{S}=\frac{1}{N}\sum_{s\in S}\frac{\mathcal{G}^{s}}{\left \|\mathcal{G}^{s}\right \|_{2}} \quad
\text{s.t.} \quad\mathcal{G}^{s}=\frac{1}{|\mathbf{x}'_s|}\sum_{x \in \mathbf{x}'_s} \nabla_\lambda \log\mathbf{G_\lambda}(x),
\label{EQ_SNFK}
\end{equation}
where $|\cdot|$ denotes the cardinality of a set. We use an average pooling since it is a natural pooling scheme for Fisher kernel rather than vector concatenation. Following the Improved Fisher Kernel framework \cite{IFK}, we finally apply power normalization and $\ell_2$-normalization to the Fisher vector $\mathcal{G}^{S}$. The overall pipeline of MPP is illustrated in Figure \ref{FIG_PIPELINE}.

\section{Analysis of Multi-scale CNN Activations}
\label{ANALYSIS_SCALE}

We compare scale characteristics between traditional local features and CNN activations. It tells us that it is not suitable to directly adopt a Fisher kernel framework to multi-scale local CNN activations for representing an image. To investigate the best way for aggregating the CNN activations into a global representation, we perform empirical studies and conclude that applying scale-wise normalization of Fisher vectors is very important.

\begin{figure}
\begin{center}
\begin{tabular}{cc}
\setlength{\tabcolsep}{1.7pt}
\hspace{-3mm}
\includegraphics[width=0.45\linewidth]{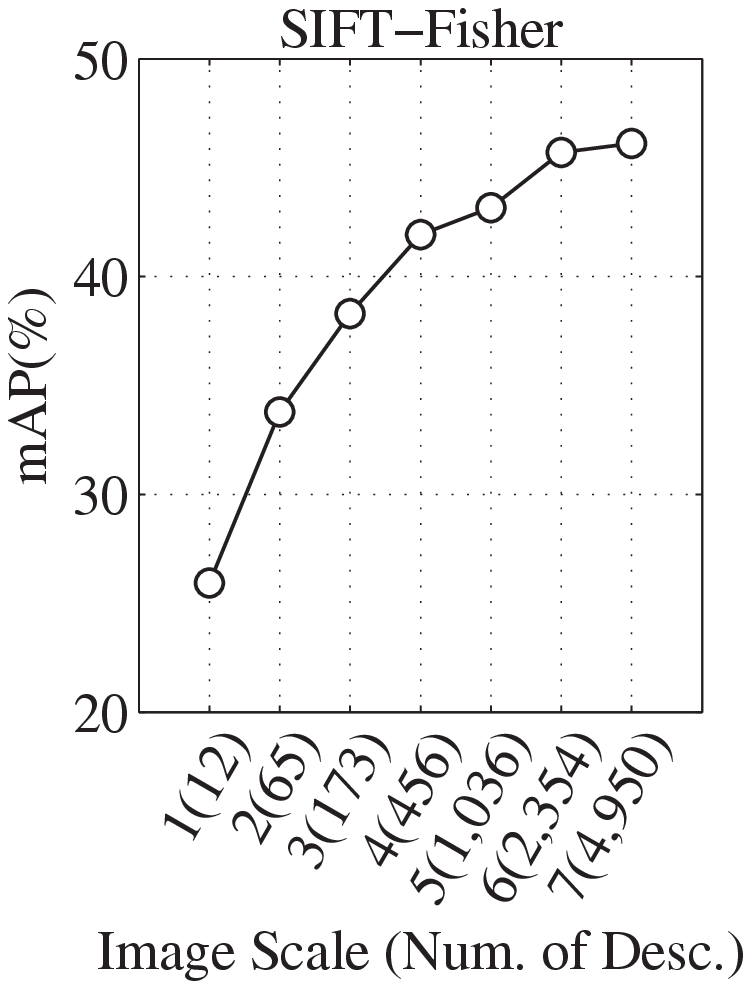}&
\hspace{-6mm}
\includegraphics[width=0.45\linewidth]{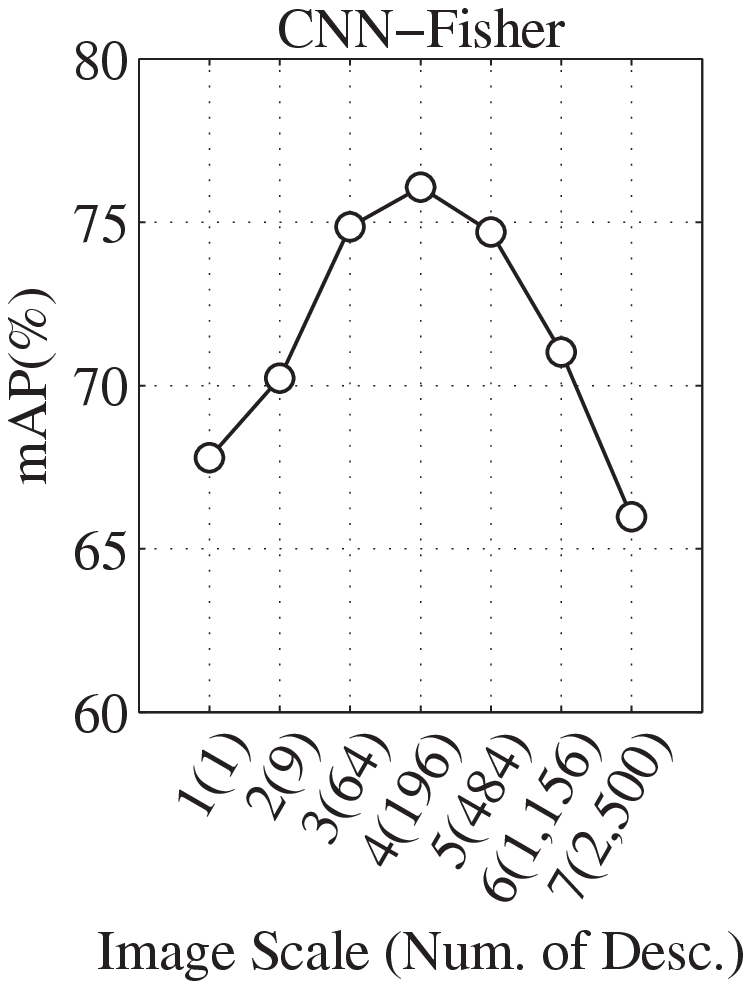}\\
\end{tabular}
\end{center}
\vspace{-6mm}
\caption{Classification performance of {SIFT-Fisher} and {CNN-Fisher} according to image scale on PASCAL VOC 2007. The tick labels of the horizontal axis denote image scales and their average number of local descriptors.}
\label{FIG_S_VS_MAP}
\end{figure}

A naive way to obtain a Fisher vector $\mathcal{G'}^{S}$ given multi-scale local activations $X=\{ x\in \mathbf{x}_s,  s \in S\}$ is to aggregate them as
\begin{equation}
\mathcal{G'}^{S}=\frac{1}{|X|}\sum_{s \in S}\sum_{x\in \mathbf{x}_{s}}\nabla_\lambda \log\mathbf{G_\lambda}(x).
\label{EQ_FK}
\end{equation}
Here, every multi-scale local activation vector is pooled to one Fisher vector with an equal weight of $1/|X|$.

To better combine a Fisher kernel with mid-level neural activations, the property of CNN activations according to patch scale should be took in consideration. In the traditional use of Fisher kernel on visual classification tasks, the hand-designed local descriptors such as SIFT \cite{SIFT} have been often densely computed in multi-scale. This local descriptor encodes low-level gradient information within an local region and captures detailed textures or shapes within a small region rather than the global structure within a larger region. In contrast, a mid-level neural activation extracted from a higher layer of CNNs (e.g. FC6 or FC7 of \cite{ALEX}) represents higher level structure information which is closer to class posteriors. As shown in the CNN visualization proposed by Zeiler and Fergus in \cite{MATTHEW}, image regions strongly activated by a certain CNN filter of the fifth layer usually capture a category-level entire object.

To figure out the different scale properties between the Fisher vector of traditional SIFT ({SIFT-Fisher}) and that of neural activation from FC7 ({CNN-Fisher}), we conduct an empirical analysis with scale-wise classification scores on PASCAL VOC 2007~\cite{VOC2007}. For the analysis, we first diversify dataset into seven different scale levels from the smallest scale of $227\times227$ resolution to the biggest scale of $1,816\times1,816$ resolution and extract both dense SIFT descriptors and local activation vectors in the seventh layer (FC7) of our CNN. Then, we follow the standard framework to encode Fisher vectors and to train an independent linear SVM for each scale, respectively.

In \Fref{FIG_S_VS_MAP}, we show the results of classification performances using {SIFT-Fisher} and {CNN-Fisher} according to scale. The figure demonstrates clear contrast between {SIFT-Fisher} and {CNN-Fisher}. {CNN-Fisher} performs worst at the largest image scale since local activations come from small image regions in an original image, while {SIFT-Fisher} performs best at the same scale since SIFT properly captures low-level contents within such small regions. If we aggregate the CNN activations of all scales into one Fisher vector by \Eref{EQ_FK}, the poorly performing 2,500 activations will have dominant influence with the large weight of 2,500/4,410 in the image representation.

\begin{figure}
\begin{center}
\begin{tabular}{ccc}
\setlength{\tabcolsep}{1.7pt}
\hspace{-10mm}
\includegraphics[width=0.38\linewidth]{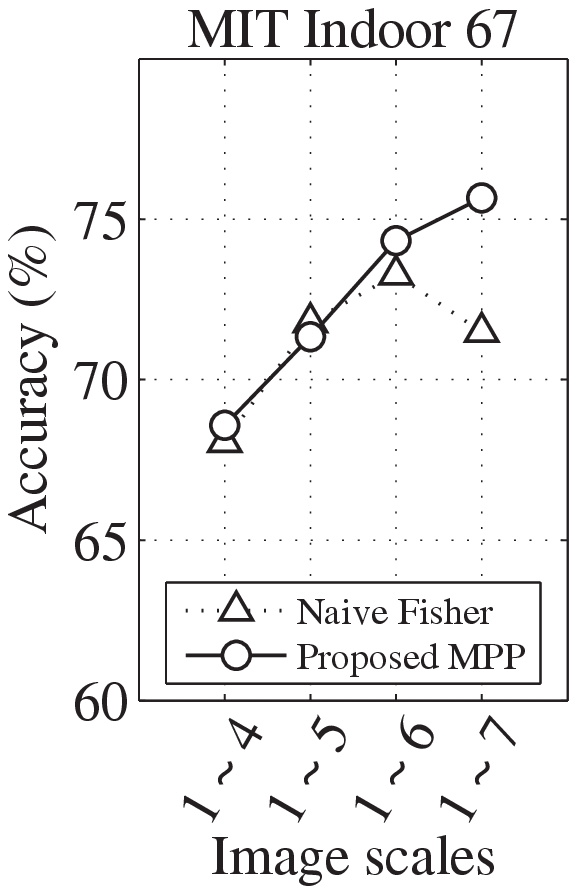}&
\hspace{-6mm}
\includegraphics[width=0.38\linewidth]{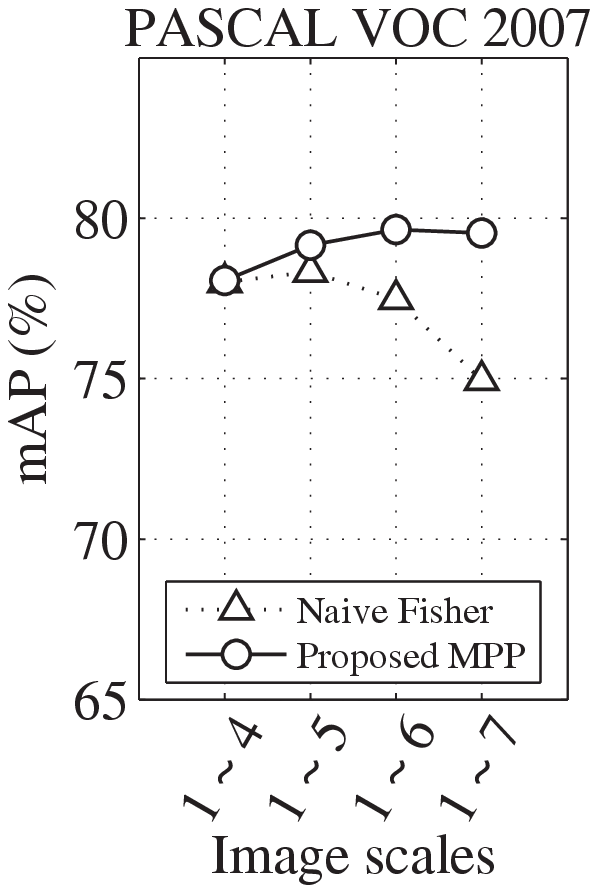}&
\hspace{-6mm}
\includegraphics[width=0.38\linewidth]{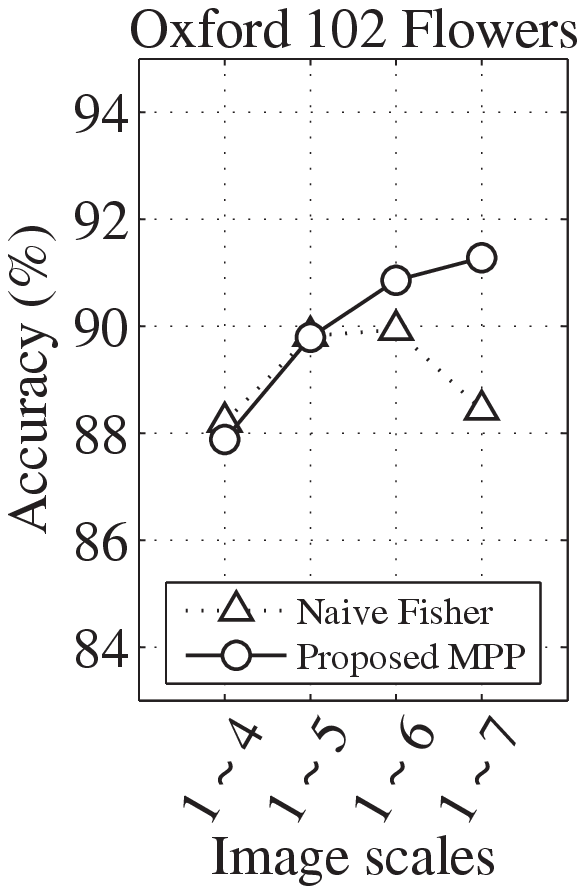}\\
\end{tabular}
\end{center}
\vspace{-4mm}
\caption{Classification performance of our \textit{multi-scale pyramid pooling} in \Eref{EQ_SNFK} and the naive Fisher pooling in \Eref{EQ_FK}. The tick labels of the horizontal axis scale levels in a scale pyramid.}
\label{FIG_SCOMBINE_VS_MAP}
\end{figure}

One possible strategy for aggregating multi-scale CNN activations is to choose activations of a set of scales relatively performing well. However, the selection of good scales is dependent on dataset and the activations from the large image scale can also contribute to geometric invariance property if we balance the influence of each scale. We empirically examined various combinations of pooling as will be shown in \Sref{sec:exp} and we found that scale-wise Fisher vector normalization followed by an simple average pooling is effective to balance the influence.

We perform an experiment to compare our pooling method in \Eref{EQ_SNFK} to the naive Fisher pooling in \Eref{EQ_FK}. In the experiment, we apply both of two pooling methods with five different numbers of scales and perform classification on PASCAL VOC 2007. Despite the simplicity of our multi-scale pyramid pooling, it demonstrates superior performances as depicted in \Fref{FIG_SCOMBINE_VS_MAP}. The performance of the naive Fisher kernel pooling in \Eref{EQ_FK} deteriorates rapidly when finer scale levels are involved. This is because indistinctive neural activations from finer scale levels become dominant in forming a Fisher vector. Our representation, however, exhibits stable performance that the accuracy is constantly increasing and finally being saturated. It verify that our pooling method aggregates multi-scale CNN activations effectively.

\section{Experiments}
\label{sec:exp}

\subsection{Datasets}
To evaluate our proposal as a generic image representation, we conduct three different visual recognition tasks with following datasets.

\paragraph{MIT Indoor 67}\cite{SCENE67} is used for a scene classification task. The dataset contains 15,620 images with 67 indoor scene classes in total. It is a challenging dataset because many indoor classes are characterized by the objects they contain (e.g. different type of stores) rather than their spatial properties. The performance is measured with top-1 accuracy.

\paragraph{PASCAL VOC 2007}\cite{VOC2007} is used for an object classification task. It consists of 9,963 images of 20 object classes in total. The task is quite difficult since the scales of the objects fluctuate and multiple objects of different classes are often contained in the same image. The performance is measured with (11-points interpolated) mean average precision.

\paragraph{Oxford 102 Flowers} \cite{FLOWERS} is used for a fine-grained object classification task, which distinguishes the sub-classes of the same object class. This dataset consists of 8,189 images with 102 flower classes. Each class consists of various numbers of images from 20 to 258. The performance is measured with top-1 accuracy.

\subsection{Pre-trained CNNs}
We use two CNNs pre-trained on the ILSVRC'12 dataset~\cite{IMAGENET} to extract multi-scale local activations. One is the Caffe reference model~\cite{CAFFE} composed of five convolutional layers and three fully connected layers. This model performed 19.6\% top-5 error when a single center-crop of each validation image are used for evaluation on the ILSVRC'12 dataset. Henceforth, we denote this model by ``{Alex}'' since it is nearly the same architecture of Krizhevsky~\etal's CNN~\cite{ALEX}. 

The other one is Chatfield~\etal's CNN-S model~\cite{DEVIL} (``{CNNS}'', henceforth). This model, a simplified version of the OverFeat~\cite{OVERFEAT}, is also composed of five convolutional layers (three in \cite{OVERFEAT}) and three fully connected layers. It shows 15.5\% top-5 error on the ILSVRC'12 dataset with the same center-crop. Compared to Alex, it uses 7$\times$7 smaller filters but dense stride of 2 in the first convolutional layer. 

Our experiments are conducted mostly with the {Alex} by default. The {CNNS} is used only for the PASCAL VOC 2007 dataset to compare our method with \cite{DEVIL}, which demonstrates excellent performance with the {CNNS}. Both of the two pre-trained models are available online \cite{MATCONVNET}.

\subsection{Implementation Details}
We use an image pyramid of seven scales by default since the seven scales can cover large enough scale variations and performance in all datasets as shown in \Fref{FIG_SCOMBINE_VS_MAP}. 

The overall procedure of our image representation is as follows. Given an image, we make an image pyramid containing seven scaled images. Each image in the pyramid has twice resolution than the previous scale starting from the standard size defined in each CNN (e.g. $227\times227$ for {Alex}). We then feed each scale image to the CNN and obtain 4,410 vectors of 4,096 dimensional dense CNN activations from the seventh layer. The dimensionality of each activation  vector is reduced to 128 by PCA where a projection is trained with 256,000 activation vectors sampled from training images. A visual vocabulary (GMM of 256 Gaussian distributions) is also trained with the same samples. Consequently, one 65,536 dimensional Fisher vector is computed by \Eref{EQ_SNFK}, and further power- and $\ell_2$-normalization follow. One-versus-rest linear SVMs with a quadratic regularizer and a hinge loss are trained finally.

Our system is mostly implemented using open source libraries including VLFeat~\cite{VLFEAT} for a Fisher kernel framework and MatConvNet~\cite{MATCONVNET} for CNNs. 

\subsection{Results and Analysis}
\label{EXPANALYSIS}
We perform comprehensive experiments to compare various methods on the three recognition tasks. We first show the performance of our method and baseline methods. Then, we compare our result with state-of-the-art methods for each dataset. For simplicity, we use a notation protocol ``A(B)'' where A denotes a pooling method and B denotes descriptors to be pooled by A. The notations are summarized in \Tref{TAB_NOTATION}.

We compare our method with several baseline methods. The baseline methods include intermediate CNN activations from a pre-trained CNN with a standard input, an average pooling with multiple jittered images, and modified versions of our method. The comparison results for each dataset are summarized in \Tref{TAB_SCENE67}(a), \ref{TAB_VOC2007}(a), \ref{TAB_FLOWERS}(a). As expected, the most basic representation, Alex-FC7, performs the worst for all datasets. The average pooling in AP10 and AP50 improves the performance +1.39\%$\sim$+3\%, however the improvement is bounded regardless of the number of data augmentation. The other two baseline methods (MPP w/o SN and CSF) exploit multi-scale CNN activations and they show better results than single-scale representations. Compared to the AP10, the performance gains from multi-scale activations exceed +10\%, +1\%, and +5\% for each dataset. It shows that image representation based on CNN activations can be enriched by utilizing multi-scale local activations.

Even though baseline methods exploiting multi-scale CNN activations show substantial improvements compared to the single-scale baselines, we can also verify that handling multi-scale activations is important for further improvement.
Compared to the naive Fisher kernel pooling (NFK) in \Eref{EQ_FK}, our MPP achieves an extra but significant performance gain of +4.18\%, +4.58\%, and 2.84\% for each dataset. Instead of pooling multi-scale activations as our MPP, concatenating encoded Fisher vectors can be another option as done in Gong~\etal's method~\cite{VLAD}. The concatenation (CSF) also improves the performance, however the CSF without an additional dimension reduction raises the dimensionality proportional to the number of scales and the MPP still outperforms the CSF for all datasets. The comprehensive test with various pooling strategies so far shows that the proposed image representation can be used as a primary image representation in wide visual recognition tasks.

We also apply the spatial pyramid (SP) kernel~\cite{SPM} to our representation. We construct a spatial pyramid into four sub-regions (whole, top, middle, bottom) and it increases the dimensionality of our representation four times. The results are unequable but the differences are marginal for all datasets. This result is not surprising because the rich activations from smaller image scales already cover the global layout. It makes the SP kernel redundant.

In \Tref{TAB_SCENE67}(b), we compare our result with various state-of-the-art methods on Indoor 67.
Similar to ours, Gong~\etal~\cite{VLAD} proposed a pooling method for multi-scale CNN activations. They performed VLAD pooling at each scale and concatenated them. Compared to \cite{VLAD}, our representation largely outperforms the method with a gain of +7.07\%. The performance gap possibly comes from 1) the large number of scales, 2) the superiority of the Fisher kernel, and 3) the details of pooling strategy. While they use only three scales, we extract seven-scale activations with a quite efficient way (\Fref{FIG_LOCALDESC}). \textit{Though adding local activations from very finer scales such as 6 or 7 in a naive way may harm the performance, it actually contribute to a better invariance property by the proposed MPP}. In addition, as our experiment of the ``CSF" was shown, the MPP is more suitable for aggregating multi-scale activations than the concatenation. It implies that our better performance does not just come from the superior Fisher kernel, but from the better handling of multi-scale neural activations.

The record holder in the Indoor 67 dataset has been Zuo~\etal~\cite{DSFL} who combined the Alex-FC6 and their complementary features so called DSFL. The DSFL learns discriminative and shareable filters with a target dataset. When we stack an additional MPP at the Pool5 layer, we already achieve a state-of-the-art performance (77.76\%) with a pre-trained Alex only. We also stack the DSFL feature\footnote{Pre-computed DSFL vectors for the MIT Indoor 67 dataset are provided by the authors.} over our representation and the result shows the performance of 80.78\%. It shows that our representation is also improved by combining complementary features. 

The results on VOC 2007 is summarized in \Tref{TAB_VOC2007}(b). There are two methods (\cite{SIVIC} and \cite{OFFTHESHELF}) that use the same Alex network. Razavian~\etal~\cite{OFFTHESHELF} performed target data augmentation and Oquab \etal \cite{SIVIC} used a multi-layer perceptron (MLP) instead of a linear SVM with ground truth bounding boxes. Our representation outperforms the two methods using the pre-trained Alex without data augmentation or the use of bounding box annotations. The gains are +1.84\% and +2.34\% respectively.

There are recent methods outperforming our method. All of them are adopting better CNNs for the source task (i.e. ImageNet classification) or the target task, such as Spatial Pyramid Pooling (SPP) network \cite{SPP}, Multi-label CNN \cite{NUS} and the CNNS \cite{DEVIL}. Our basic MPP(Alex-FC7) demonstrates slightly lower precisions  (79.54\%) compared to them, however we use the basic Alex CNN without fine-tuning on VOC 2007. When our representation is equipped with the superior CNNS \cite{DEVIL}, which is not fine-tuned on VOC 2007, our representation (81.40\%) reaches nearly stat-of-the-art performance and our method is further improved to 82.13\% by stacking MPP(CNNS-FC8). The performance is still lower than \cite{DEVIL}, who conduct target data augmentation and fine-tuning. We believe our method can be further improved by additional techniques such as fine-tuning, target data augmentation, or use of ground truth bounding boxes, we leave the issue as future work because our major focus is a generic image representation with a pre-trained CNN.

\Tref{TAB_FLOWERS}(b) shows the classification performances on 102 Flowers. Our method (91.28\%) outperforms the previous state-of-the-art method~\cite{OFFTHESHELF} (86.80\%). Without the use of a powerful CNN representation, various previous methods show much lower performances.

\Tref{TAB_VOC2007_PERCLASS} shows per-class performances on VOC 2007. Compared to state-of-the-art methods, our method performs best in 6 classes among 20 classes. It is interesting that the 6 classes include ``bottle", ``pottedplant", and ``tvmonitor", which are the representative small objects in the VOC 2007 dataset. The results clearly demonstrates the benefit of our MPP that aggregates activations from very finer-scales as well, which are prone to harm the performance if it is handled inappropriately.

\begin{table*}
\setlength{\tabcolsep}{1.6pt}
\small
\begin{center}
\begin{tabular}{|l|l|}
\hline
Method&Description\\
\hline\hline
CNN-FC7& A standard activation vector from FC7 of a CNN with a center-crop of a $256\times256$ size input image.\\\hline
AP10(CNN-FC7)&Average pooling of a 5 crops and their flips, given a $256\times256$ size input image.\\\hline
AP50(CNN-FC7)&Average pooling of a 25 crops and their flips, given a $256\times256$ size input image.\\\hline
NFK(CNN-FC7)&Naive Fisher kernel pooling without scale-wise vector normalization, given a multi-scale image pyramid.\\\hline
CSF(CNN-FC7)&Concatenation of scale-wise normalized Fisher vectors, given a multi-scale image pyramid.\\\hline
MPP(CNN-FC7)&The proposed representation, given a multi-scale image pyramid.\\\hline
\end{tabular}
\end{center}
\caption{Summary of our notation protocol. Consequent image representations by the listed methods are finally $\ell_2$-normalized.}
\label{TAB_NOTATION}
\end{table*}

\begin{table}
\setlength{\tabcolsep}{1.6pt}
\small
\begin{center}
\begin{tabular}{|l|l|l|c|}\hline
Method		&Description								&CNN&Acc.\\\hline\hline
Baseline	&Alex-FC7									&Yes.&57.91\\
Baseline	&AP10(Alex-FC7)							&Yes.&60.90\\
Baseline	&AP50(Alex-FC7)							&Yes.&60.37\\
Baseline	&NFK(Alex-FC7)					&Yes.&71.49\\
Baseline	&CSF(Alex-FC7)			&Yes.&72.24\\
Ours		&MPP(Alex-FC7)							&Yes.&75.67\\
Ours		&MPP(Alex-FC7)+SP 						&Yes.&\textbf{75.97}\\\hline\hline
Ours		&MPP(Alex-FC7,Pool5)&Yes.&77.56\\
Ours		&MPP(Alex-FC7)+DSFL\cite{DSFL}	&Yes.&\textbf{80.78}\\\hline
\end{tabular} \\(a) baselines and our methods.\\
\begin{tabular}{|l|l|l|c|}\hline
Method&Description&CNN&Acc.\\\hline\hline
Singh \etal \cite{UNSUPERMID} '12&Part+GIST+DPM+SP&No.&49.40\\
Juneja \etal \cite{SHOUT} '13&IFK+Bag-of-Parts&No.&63.18\\
Doersch \etal \cite{MIDREP} '13&IFK+MidlevelRepresent.&No.&66.87\\
Zuo \etal \cite{DSFL} '14&DSFL&No.&52.24\\
Zuo \etal \cite{DSFL} '14&DSFL+Alex-FC6&Yes.&\textbf{76.23}\\
Zhou \etal \cite{MITPLACE} '14&Alex-FC7&Yes.&68.24\\
Zhou \etal \cite{MITPLACE} '14&Alex-FC7&Yes.&70.80\\
Razavian \etal~\cite{OFFTHESHELF} '14&AP(Alex)+PT+TargetAug.&Yes.&69.00\\
Gong \etal~\cite{VLAD} '14&VLAD Concat.(Alex-FC7)&Yes.&68.90\\\hline
\end{tabular} \\(b) state-of-the-art methods on MIT Indoor 67.\\
\end{center}
\caption{Classification performances on MIT Indoor 67. (SP: Spatial Pyramid, DPM: Deformable Part-based Model, PT: Power Transform, IFK: Improved Fisher Kernel, DSFL: Discriminative and Shareable Feature Learning.)}
\vspace{20mm}
\label{TAB_SCENE67}
\end{table}

\begin{table}
\setlength{\tabcolsep}{1.0pt}
\small
\begin{center}
\begin{tabular}{|l|l|l|l|l|c|}\hline
Method		&Description						&FT&BB	&CNN&mAP\\\hline\hline
Baseline	&Alex-FC7							&No.			&No. 		&Yes.&72.36\\
Baseline	&AP10(Alex-FC7)					&No.			&No. 		&Yes.&73.75\\
Baseline	&AP50(Alex-FC7)					&No.			&No. 		&Yes.&73.60\\
Baseline	&NFK(Alex-FC7)			&No.			&No. 		&Yes.&74.96\\
Baseline	&CSF(Alex-FC7)	&No.			&No. 		&Yes.&78.46\\
Ours		&MPP(Alex-FC7)					&No.			&No. 		&Yes.&\textbf{79.54}\\
Ours		&MPP(Alex-FC7)+SP 				&No.			&No. 		&Yes.&79.29\\\hline\hline
Ours		&MPP(CNNS-FC7) 					&No.			&No. 		&Yes.&81.40\\
Ours		&MPP(CNNS-FC7,FC8)&No.			&No. 		&Yes.&\textbf{82.13}\\\hline
\end{tabular} \\(a) Baselines and our methods.\\
\begin{tabular}{|l|l|l|l|l|c|}\hline
Method		&Description						&FT&BB	&CNN&mAP\\\hline\hline
\bigcell{l}{Perronnin \etal~\cite{IFK}\;10'}&IFK(SIFT+color)&No.&No.&No.&60.3\%\\
\bigcell{l}{He \etal~\cite{SPP}\;'14}&SPPNET-FC7&No.&No.&Yes.&80.10\%\\
\bigcell{l}{Wei \etal~\cite{NUS}\;'14}&Multi-label CNN&Yes.&No.&Yes.&81.50\%\\
Razavian \etal~\cite{OFFTHESHELF}\;'14&AP(Alex)+PT+TA&No.&No.&Yes.&77.20\%\\
Oquab \etal~\cite{SIVIC}\;'14&Alex-FC7+MLP&No.&Yes.&Yes.&77.70\%\\
\bigcell{l}{Chatfield~\etal \cite{DEVIL}\;'14}&\bigcell{l}{AP(CNNS-FC7)+TA}&No.&No.&Yes.&79.74\%\\
\bigcell{l}{Chatfield~\etal \cite{DEVIL}\;'14}&\bigcell{l}{AP(CNNS-FC7)+TA}&Yes.&No.&Yes.&\textbf{82.42}\%\\\hline
\end{tabular} \\(b) state-of-the-art methods on \\PASCAL VOC 2007 classification.\\
\end{center}
\vspace{-3mm}
\caption{Classification performances on PASCAL VOC 2007 classification. ``FT" represents fine-tuning of a pre-trained CNN on VOC2007 and ``BB'' denotes the use of ground truth object bounding boxes in training. (SP: Spatial Pyramid, IFK: Improved Fisher Kernel, SPPNET: Spatial Pyramid Pooling Network, PT: Power Transform, TA: Target data Augmentation in training, MLP: Multilayer Perceptron.)}
\vspace{25mm}
\label{TAB_VOC2007}
\end{table}

\begin{table*}
\setlength{\tabcolsep}{1.4pt}
\footnotesize
\begin{center}
\begin{tabular}{|l|c||c|c|c|c|c|c|c|c|c|c|c|c|c|c|c|c|c|c|c|c|c|}\hline
Method		&FT& plane & bike  & bird  & boat  & bottle & bus   & car   & cat   & chair & cow   & table & dog   & horse & motor & person & plant & sheep & sofa  & train & tv    & mAP \\\hline\hline
Alex-FC7&No.&85.0& 	79.7&	82.8&	80.4&	39.7&	69.3&	82.9&	81.7&	58.7&	57.8&	68.5&	75.9&	83.0&	72.5&	90.6&	51.7&	71.1&	60.8&	85.0&	70.5&	72.4\\
AP10(Alex-FC7)&No.&85.7&  	80.8&	83.3&	80.7&	40.4&	71.5&	83.8&	82.7&	60.7&	60.5&	70.6&	79.0&	84.5&	75.0&	91.3&	53.4&	70.1&	62.6&	86.5&	72.1&	73.7\\
MPP(Alex-FC7)&No.&90.2&  	86.9&	86.6&	84.4&	54.0&	80.0&	87.9&	86.0&	63.4&	72.2&	75.7&	83.1&	87.8&	83.9&	93.0&	\textbf{64.8}&	75.8&	69.6&	89.9&	75.9&	79.5\\
MPP(CNNS-FC7)&No.		&90.2 &  	88.6 &	89.0 &	84.7 &	\textbf{58.2}&\textbf{82.8} &	88.1 &	89.0 &	\textbf{64.9} &	77.0 &	\textbf{78.4} &	86.9 &	89.2 &	86.7 &	92.8 &	61.2 &	81.3 &	\textbf{70.0} &	89.8 &	\textbf{79.3} &	\textbf{81.4}\\\hline\hline
Perronnin \etal\cite{IFK}\;10'&No.&75.7&   64.8&   52.8&   70.6&   30.0&   64.1&   77.5&   55.5&   55.6&   41.8&   56.3&   41.7&   76.3&   64.4&   82.7&   28.3&   39.7&   56.6&   79.7&   51.5&   58.3\\
Razavian \etal\cite{OFFTHESHELF}\;'14&No.&90.1&	84.4&   86.5&   84.1&   48.4&   73.4&   86.7&   85.4&   61.3&   67.6&   69.6&   84.0&   85.4&   80.0&   92.0&   56.9&   76.7&   67.3&   89.1&   74.9&   77.2\\
Oquab \etal\cite{SIVIC}\;'14&No.&88.5&	81.5&	87.9&	82.0&	47.5&	75.5&	90.1&	87.2&	61.6&	75.7&	67.3&	85.5&	83.5&	80.0&	95.6&	60.8&	76.8&	58.0&	90.4&	77.9&	77.7\\
Wei \etal\cite{NUS}\;'14&Yes.&95.1&	    90.1&\textbf{92.8}&	\textbf{89.9}&	51.5&	80.0&	\textbf{91.7}&	91.6&	57.7&	\textbf{77.8} &	70.9&	89.3&	89.3&	85.2&	93.0&	64.0&	\textbf{85.7}&	62.7&	94.4&	78.3&	81.5\\
Chatfield\etal \cite{DEVIL}\;'14&Yes.&\textbf{95.3}&\textbf{90.4}&	92.5&	89.6&	54.4&	81.9&	91.5&	\textbf{91.9}&	64.1&	76.3&   74.9&	\textbf{89.7}&	\textbf{92.2}&	\textbf{86.9}&	\textbf{95.2}&	60.7&	82.9&	68.0&	\textbf{95.5}&	74.4&	\textbf{82.4}\\\hline
\end{tabular} 
\end{center}
\caption{Per-class classification performances on PASCAL VOC 2007. ``FT" represents fine-tuning of a pre-trained CNN on VOC2007.}
\label{TAB_VOC2007_PERCLASS}
\end{table*}

\begin{table*}[!h]
\setlength{\tabcolsep}{1.0pt}
\small
\begin{center}
\begin{tabular}{|l|l|l|l|l|c|}\hline
Method		&Description						&Seg.&CNN&Acc.\\\hline\hline
Baseline	&Alex-FC7							&No.&Yes.&81.43\\
Baseline	&AP10(Alex-FC7)					&No.&Yes.&83.40\\
Baseline	&AP50(Alex-FC7)					&No.&Yes.&83.56\\
Baseline	&NFK(Alex-FC7)			&No.&Yes.&88.44\\
Baseline	&CSF(Alex-FC7)	&No.&Yes.&89.35\\
Ours		&MPP(Alex-FC7)					&No.&Yes.&\textbf{91.28}\\
Ours		&MPP(Alex-FC7)+SP 				&No.&Yes.&90.05\\\hline
\end{tabular} \\(a) baselines and our methods.\\
\begin{tabular}{|l|l|l|l|l|c|}\hline
Method		&Description						&Seg.&CNN&Acc.\\\hline\hline
\bigcell{l}{Nilsback and Zisserman \cite{FLOWERS}\;'08}&Multple kernel learning&Yes.&No.&77.70\\
\bigcell{l}{Angelova and Zhu \cite{SEGCLS}\;'13}&\bigcell{l}{Seg+DenseHoG+LLC+MaxPooling}&Yes.&No.&80.70\\
\bigcell{l}{Murray and Perronnin \cite{GMP}\;'14}&GMP of IFK(SIFT+color)&No.&No.&81.50\\
\bigcell{l}{Fernando \etal~\cite{FLH}\;'14}&Bag-of-FLH&Yes.&No.&72.70\\
Razavian \etal~\cite{OFFTHESHELF}\;'14&AP(Alex)+PT+TA&No.&Yes.&\textbf{86.8}\\\hline
\end{tabular} \\(b) state-of-the-art methods on Oxford 102 Flowers.\\
\end{center}
\vspace{-3mm}
\caption{Classification performances on Oxford 102 Flowers. ``Seg.'' denotes the use of ground truth segmentations in training. (SP: spatial pyramid, LLC: Locality-constrained Linear Coding, GMP: generalized max pooling, FLH: Frequent Local Histograms, PT: power transform, TA: target data augmentation in training.)}
\label{TAB_FLOWERS}
\end{table*}

\begin{figure*}[!hb]
\begin{center}
\small
\scalebox{0.9}{\begin{tabular}{cccc}
\setlength{\tabcolsep}{1.7pt}
\includegraphics[width=0.23\linewidth]{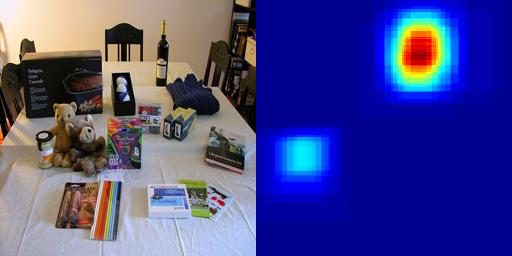}&
\includegraphics[width=0.23\linewidth]{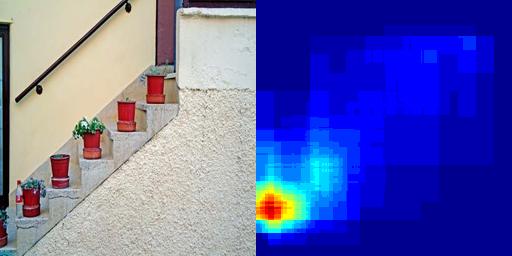}&
\includegraphics[width=0.23\linewidth]{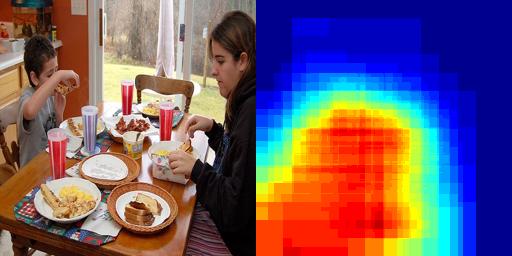}&
\includegraphics[width=0.23\linewidth]{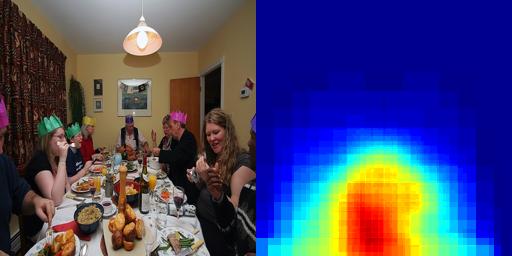}\\
Bottle&Bottle&Dining Table&Dining Table\\
\includegraphics[width=0.23\linewidth]{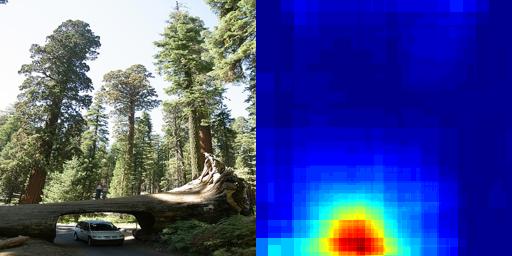}&
\includegraphics[width=0.23\linewidth]{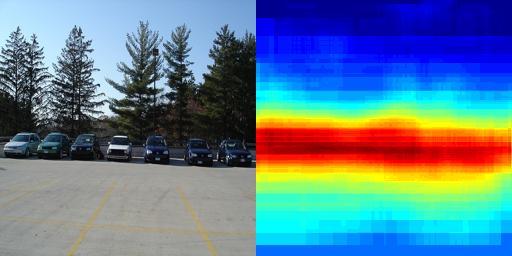}&
\includegraphics[width=0.23\linewidth]{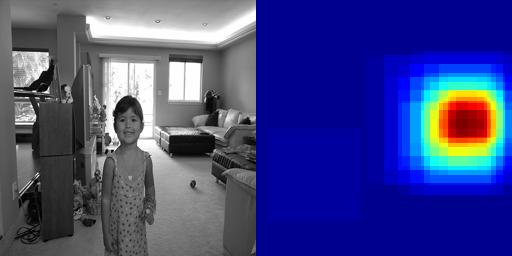}&
\includegraphics[width=0.23\linewidth]{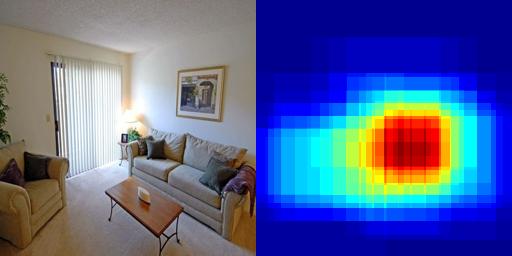}\\
Car&Car&Sofa&Sofa\\
\includegraphics[width=0.23\linewidth]{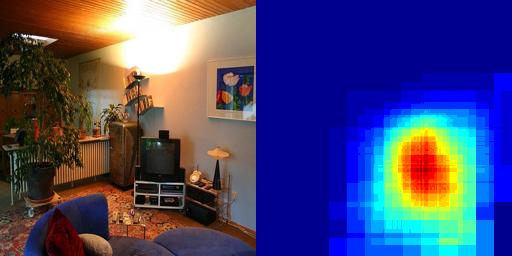}&
\includegraphics[width=0.23\linewidth]{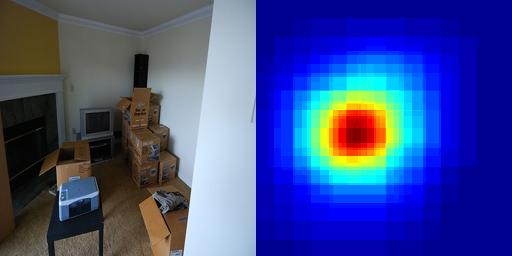}&
\includegraphics[width=0.23\linewidth]{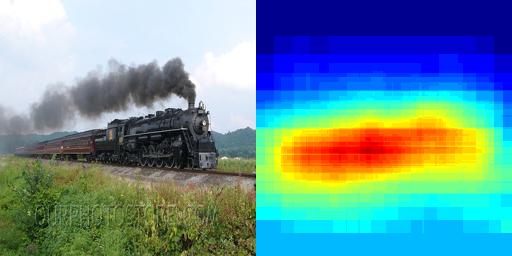}&
\includegraphics[width=0.23\linewidth]{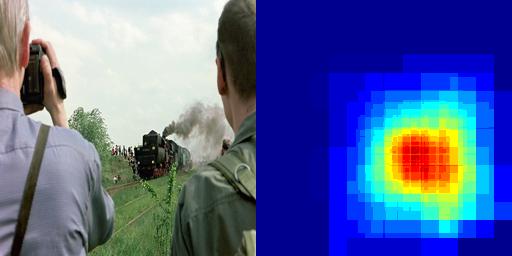}\\
TV monitor&TV monitor&Train&Train\\
\includegraphics[width=0.23\linewidth]{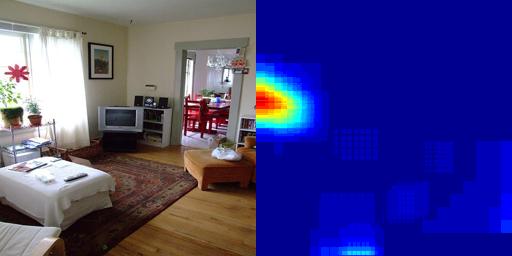}&
\includegraphics[width=0.23\linewidth]{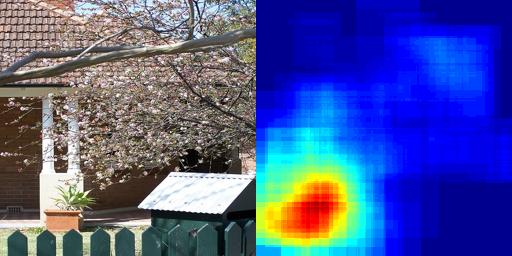}&
\includegraphics[width=0.23\linewidth]{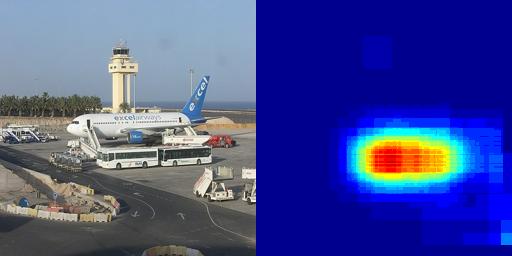}&
\includegraphics[width=0.23\linewidth]{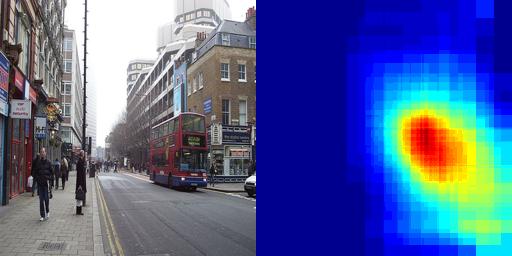}\\
Potted plant&Potted plant&Bus&Bus\\
\includegraphics[width=0.23\linewidth]{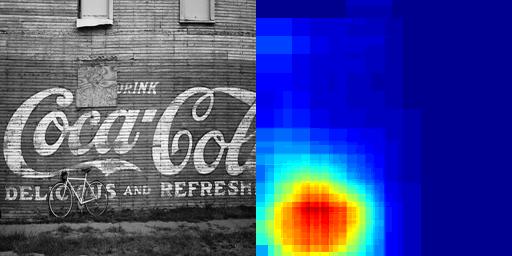}&
\includegraphics[width=0.23\linewidth]{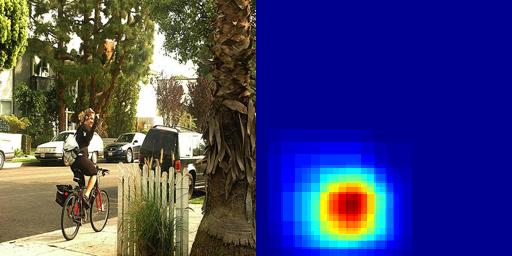}&
\includegraphics[width=0.23\linewidth]{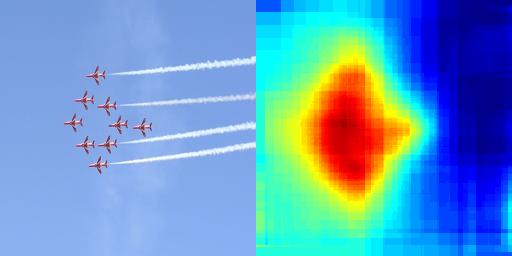}&
\includegraphics[width=0.23\linewidth]{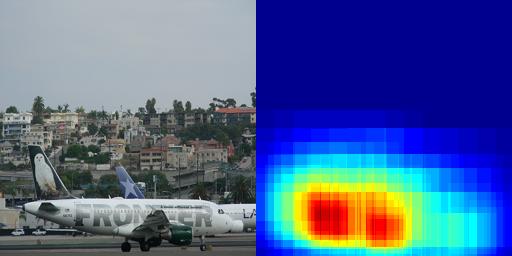}\\
Bicycle&Bicycle&Aeroplane&Aeroplane\\
\includegraphics[width=0.23\linewidth]{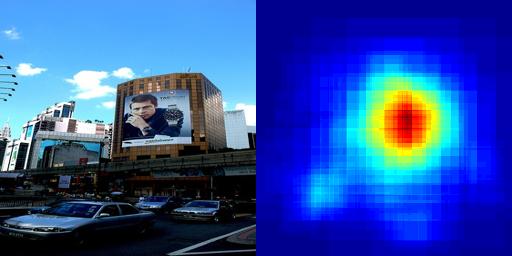}&
\includegraphics[width=0.23\linewidth]{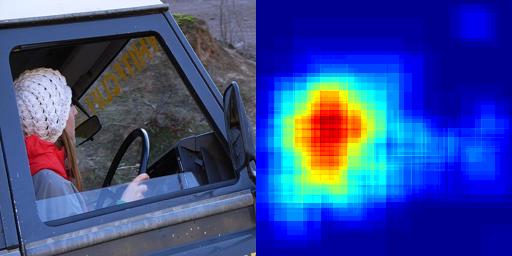}&
\includegraphics[width=0.23\linewidth]{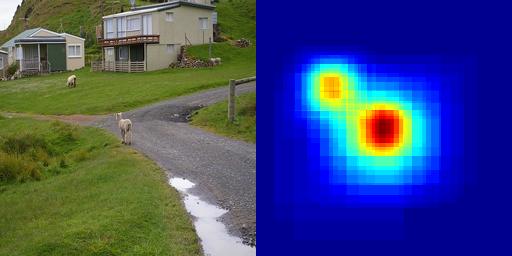}&
\includegraphics[width=0.23\linewidth]{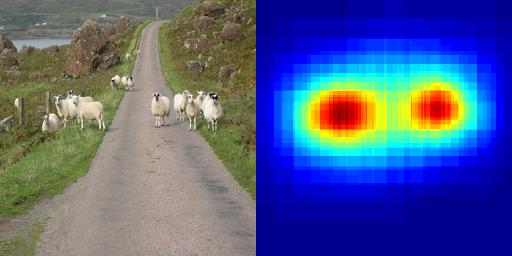}\\
Person&Person&Sheep&Sheep\\
\end{tabular}}
\end{center}
\vspace{-3mm}
\caption{Examples of object confidence maps obtained by our image representation on the PASCAL VOC 2007. All examples are test images, not training images.}
\label{FIG_OBJCONF}
\end{figure*}
\FloatBarrier
\clearpage\clearpage\clearpage\clearpage

\subsection{Weakly-Supervised Object Confidence Map}
\begin{figure}
\small
\begin{center}
\begin{tabular}{cc}
\includegraphics[width=0.45\linewidth]{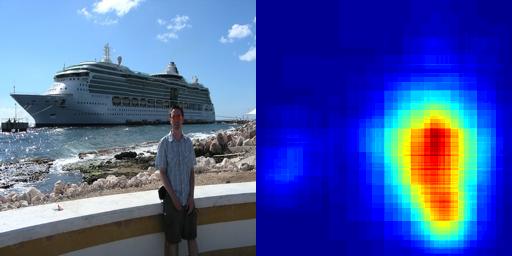}&
\includegraphics[width=0.45\linewidth]{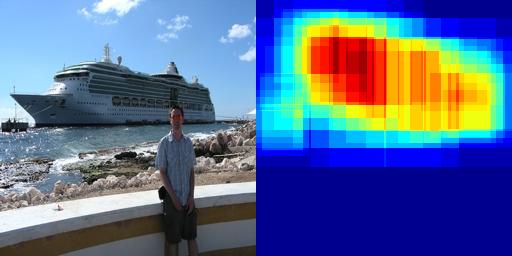}\\
Person&Boat\\
\includegraphics[width=0.45\linewidth]{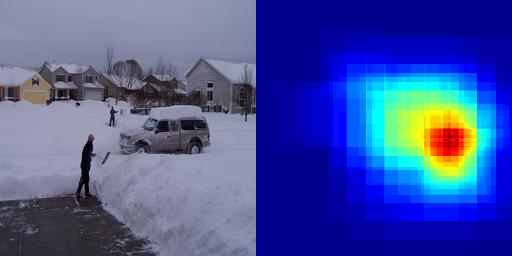}&
\includegraphics[width=0.45\linewidth]{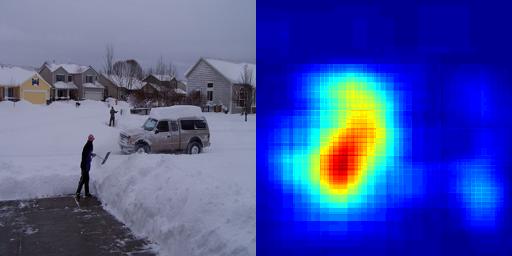}\\
Car&Person\\
\includegraphics[width=0.45\linewidth]{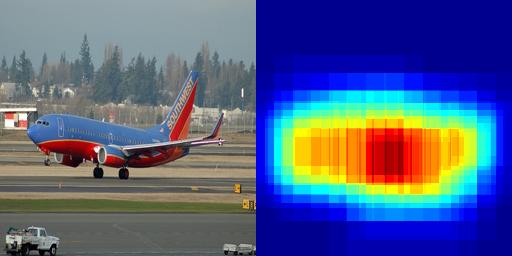}&
\includegraphics[width=0.45\linewidth]{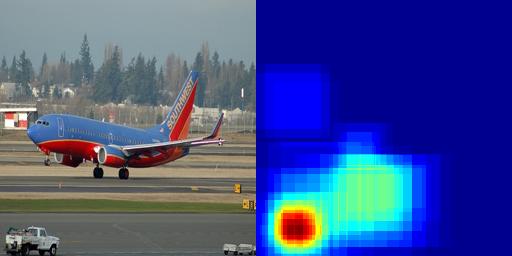}\\
Aeroplane&Car\\
\includegraphics[width=0.45\linewidth]{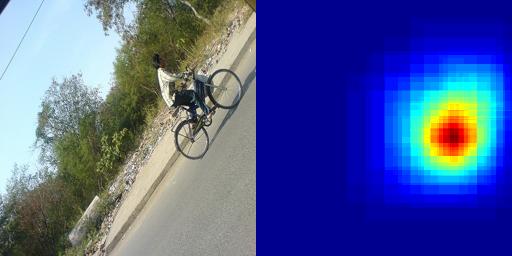}&
\includegraphics[width=0.45\linewidth]{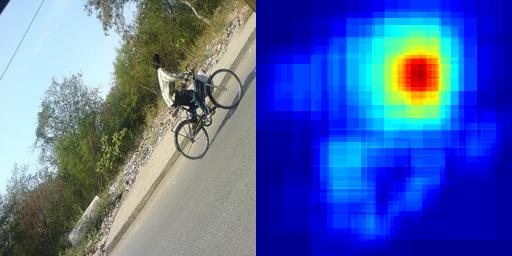}\\
Bicycle&Person\\
\includegraphics[width=0.45\linewidth]{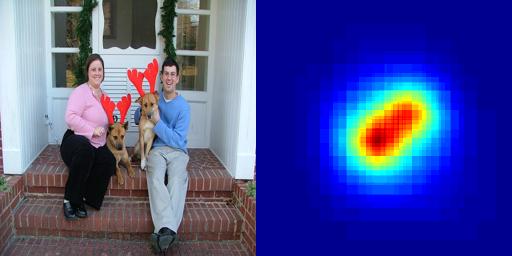}&
\includegraphics[width=0.45\linewidth]{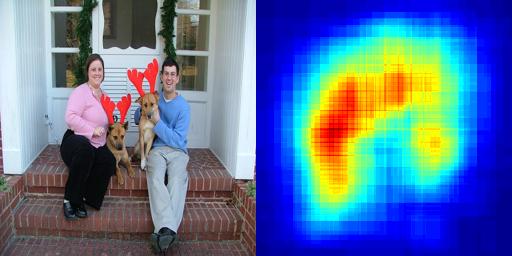}\\
Dog&Person\\
\includegraphics[width=0.45\linewidth]{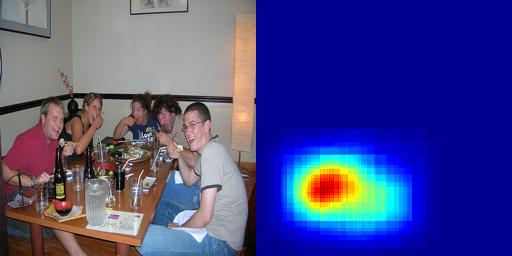}&
\includegraphics[width=0.45\linewidth]{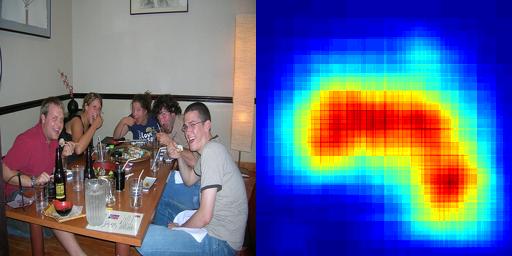}\\
Bottle&Person\\
\includegraphics[width=0.45\linewidth]{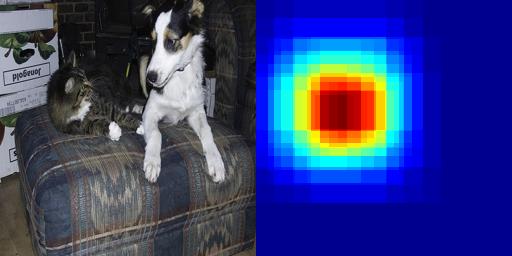}&
\includegraphics[width=0.45\linewidth]{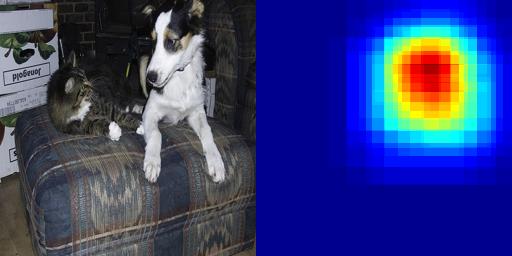}\\
Cat&Dog\\
\includegraphics[width=0.45\linewidth]{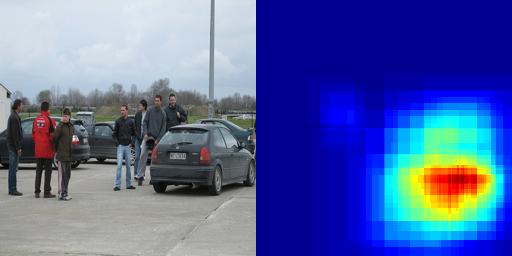}&
\includegraphics[width=0.45\linewidth]{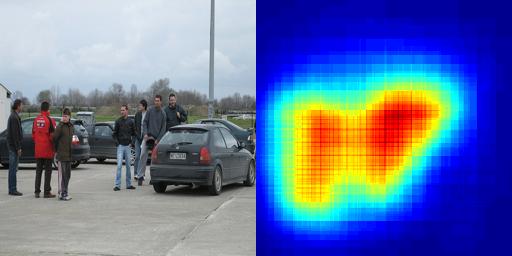}\\
Car&Person\\
\includegraphics[width=0.45\linewidth]{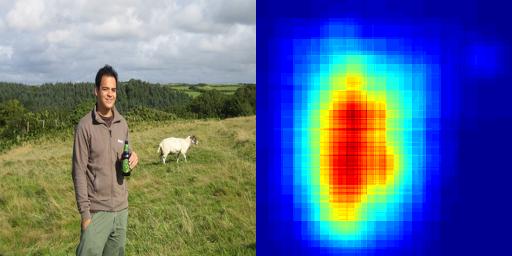}&
\includegraphics[width=0.45\linewidth]{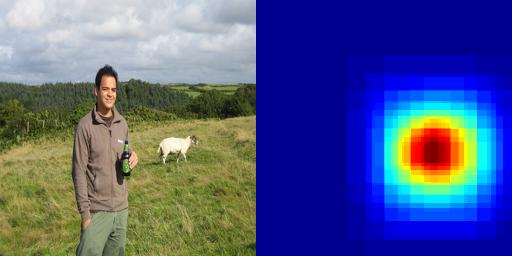}\\
Person&Sheep\\
\end{tabular}
\end{center}
\caption{Examples of multi-object confidence map obtained by our image representation on the PASCAL VOC 2007. All examples are test images, not training images.}
\label{FIG_OBJCONF_MULTI}
\end{figure}

One interesting feature of our method is that we can present object confidence maps for object classification tasks, though we train the SVM classifiers \textit{without bounding box annotation but only with class-level labels}.
To recover confidence maps, we trace how much weight is given to each local patch and accumulate all the weights of local activations.
Tracing the weight of local activations is possible because our final representation can be formed regardless of the number of scales and the number of local activation vectors. To trace the weight of each patch, we compute our final representation per patch using the corresponding single activation vector only and compute the score from the pre-trained SVM classifiers we used for object classification.

Fig. \ref{FIG_OBJCONF} and Fig. \ref{FIG_OBJCONF_MULTI} show several examples of object confidence map on the VOC 2007 test images. In the figures, we can verify our image representation encodes the discriminative image patches well, despite large within-class variations as well as substantial geometric changes. As we discussed in \Sref{EXPANALYSIS}, the images containing small-size objects also present the accurate confidence maps. These maps may further be utilized as an considerable cue for object detection/localization and also be useful for analyzing image representation.

\section{Discussion}
We have proposed the multi-scale pyramid pooling for better use of neural activations from a pre-trained CNN. There are several conclusions we can derive through our study. First, we should take the scale characteristic of neural activations into consideration for the successful combination of a Fisher kernel and a CNN. The activations become uninformative as a patch size becomes smaller, however they can contribute to better scale invariance when they meet a simple scale-wise normalization. Second, dense deep neural activations from multiple scale levels are extracted with reasonable computations by replacing the fully connection with equivalent multiple convolution filters. It enables us to pool the truly multi-scale activations and to achieve significant performance improvements on the visual recognition tasks. Third, reasonable object-level confidence maps can be obtained from our image representation even though only class-level labels are given for supervision, which can be further applied to object detection or localization tasks. In the comprehensive experiments on three different recognition tasks, the results suggest that our proposal can be used as a primary image representation for better performances in various visual recognition tasks.

{\small
\bibliographystyle{ieee}
\bibliography{egbib}
}

\end{document}